\documentclass{article}

\usepackage{PRIMEarxiv}

\usepackage[utf8]{inputenc} % allow utf-8 input
\usepackage[T1]{fontenc}    % use 8-bit T1 fonts
\usepackage{hyperref}       % hyperlinks
\usepackage{url}            % simple URL typesetting
\usepackage{booktabs}       % professional-quality tables
\usepackage{amsfonts}       % blackboard math symbols
\usepackage{nicefrac}       % compact symbols for 1/2, etc.
\usepackage{microtype}      % microtypography
\usepackage{lipsum}
\usepackage{fancyhdr}       % header
\usepackage{graphicx}       % graphics
\graphicspath{{media/}}     % organize your images and other figures under media/ folder

\title{SyncDreamer for 3D Reconstruction of Endangered Animal Species with NeRF and NeuS
}

\author{
  Ahmet Haydar Ornek, Deniz Sen, Esmanur Civil \\
  Huawei Türkiye R\&D Center \\
  İstanbul, Türkiye\\
  \texttt{\{ahmet.haydar.ornek2, deniz.sen1, esmanur.civil1\}@huawei.com} \\
}

\begin{document}
\maketitle

\begin{abstract}

%Deadline: 14 December 2023

The main aim of this study is to demonstrate how innovative view synthesis and 3D reconstruction techniques can be used to create models of endangered species using monocular RGB images. To achieve this, we employed SyncDreamer to produce unique perspectives and NeuS and NeRF to reconstruct 3D representations. We chose four different animals, including the oriental stork, frog, dragonfly, and tiger, as our subjects for this study. Our results show that the combination of SyncDreamer, NeRF, and NeuS techniques can successfully create 3D models of endangered animals. However, we also observed that NeuS produced blurry images, while NeRF generated sharper but noisier images. This study highlights the potential of modeling endangered animals and offers a new direction for future research in this field. By showcasing the effectiveness of these advanced techniques, we hope to encourage further exploration and development of techniques for preserving and studying endangered species.

\end{abstract}

% keywords can be removed
\keywords{SyncDreamer \and NeuS \and NeRF \and 3D \and Reconstruction \and Novel View Synthesis}

\section{Introduction}

In recent years, the pervasive influence of Artificial Intelligence (AI) and deep learning technologies has substantially reshaped our daily lives, revolutionizing diverse sectors ranging from healthcare to entertainment. Within this transformative landscape, generative AI, with its ability to autonomously create realistic and intricate content, has emerged as a focal point of technological innovation. This surge in interest has been notably propelled by the advent of the stable diffusion(SD) algorithm, a groundbreaking development that has propelled generative AI to new heights. Concurrently, the field has witnessed remarkable progress in 3D generation capabilities, further expanding the potential applications of AI in various domains. As the amount of data and the advancements in the field of neural implicit 3D representations have remarkably increased, the utilization of generative AI has started to be used in the industry. However, training consistent 3D representations requires large numbers of data samples, which is not always available in certain cases.

Preserving biodiversity remains a critical concern, with many species facing the threat of extinction in the wild. Compounding this challenge is the scarcity of image data for certain endangered species, making it difficult to create 3D models using the existing generative AI-based approaches. This paper explores the intersection of 3D generative AI and wildlife conservation, discussing the results of the existing zero-shot 3D model generation approaches to address the data scarcity issue. By leveraging advanced AI techniques, particularly in generative novel view synthesis and neural implicit representations, we aim to generate 3D models of endangered species from limited existing samples. This study not only underscores the multifaceted impact of AI on real-world challenges but also exemplifies its potential to contribute meaningfully to the preservation of our planet's biodiversity. 

\begin{figure*}[h!]
    \centering
    \includegraphics[scale=0.5]{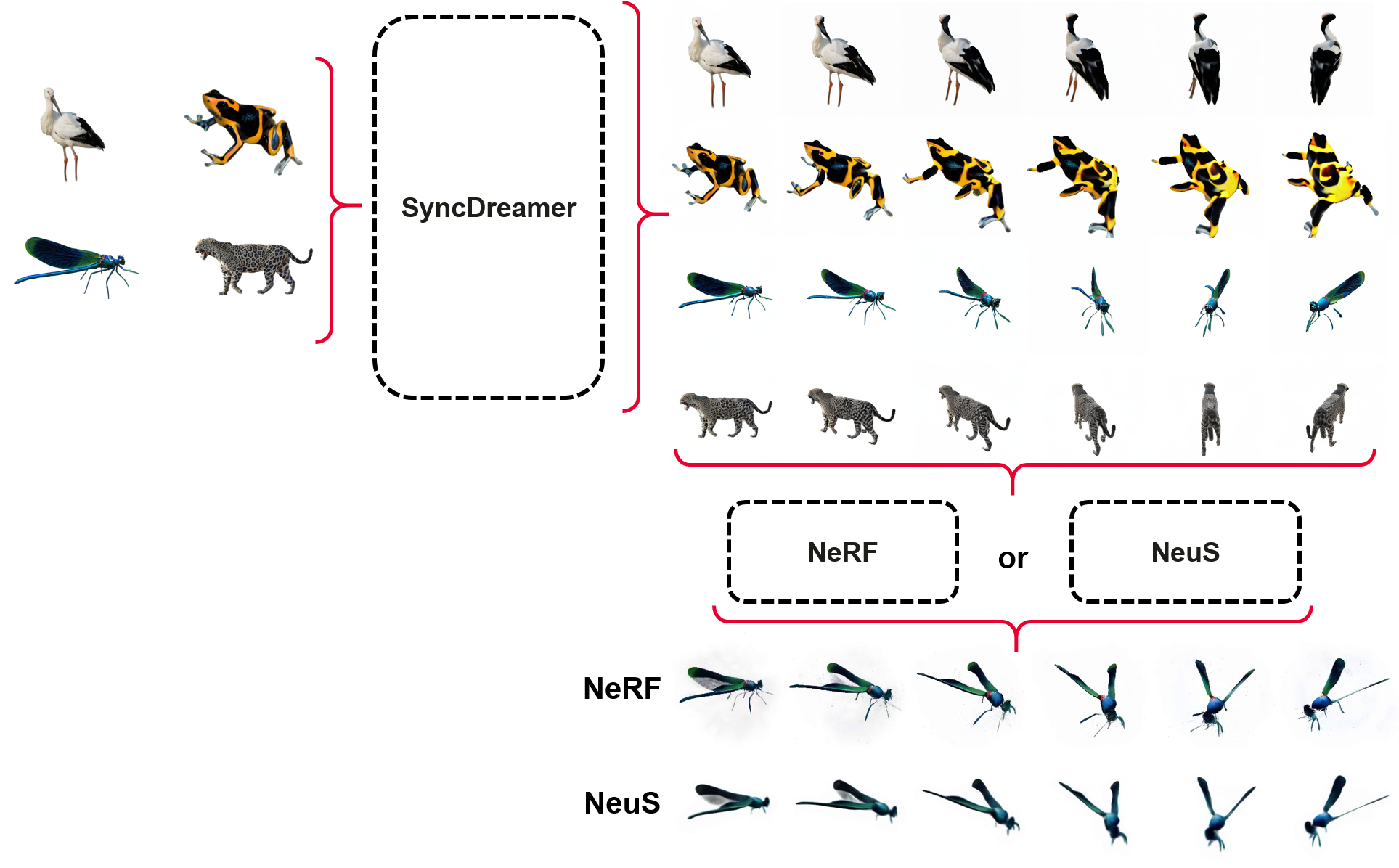}
    \caption{SyncDreamer takes monocular RGB images as input and generates their novel views in different angles. NeRF or NeuS are used to generate 3D models using the generated novel views.}
    \label{fig:graphical_abstract}
\end{figure*}

The graphical abstract of the study can be seen in Fig \ref{fig:graphical_abstract}. Our main contributions are as follows: 

\begin{itemize}
    \item Generating a novel view of endangered species using SyncDreamer
    \item Creating 3D representations using NeRF
    \item Generating 3D representations using NeuS
\end{itemize}

The paper is structured as follows: The Introduction section provides an overview of the problem and the approach taken. The Related Work section discusses previous research in the field of 3D model generation. In the Material and Method section, the datasets and tools used in the experiments are described, including Material and SyncDreamer. The proposed method is compared with NeRF and NeuS in the NeuS and NeRF section. The Experiments and Results section presents both quantitative and qualitative results. The Discussion section covers the limitations and future directions of the research. Finally, the paper concludes in the Conclusion section.

\section{Related Work} \label{sec:related_work}

This section details the Stable Diffusion (SD), Neural 3D Representations, and Single Image 3D Model Generation works.

\subsection{Stable Diffusion}
Similar to the generative adversarial networks (GANs) \cite{goodfellow2020generative}, the main aim of the stable diffusion algorithm is to map the Gaussian distribution into an image distribution; on the other hand, unlike GANs and techniques such as variational autoencoders(VAEs) which directly yield the decoded version of the sampled noise, SD models iteratively remove the noise by passing the latent image outputs of the process through a U-Net based denoising autoencoder. As a result, the optimized denoising process yields more consistent and realistic images, while not suffering from inherent problems such as mode collapse. Denoising diffusion probabilistic models(DDPMs), a foundational method, where the model is taught to reverse a forward Markov chain where the input image is iteratively perturbed with noise sampled from Gaussian distribution with a sufficiently high standard deviation \cite{ho2020denoising}. Therefore we obtain a forward and a backward Markov chain to map a noise distribution into actual image data. To generate new data, we sample a noise vector from the Gaussian distribution, similar to the single-step counterparts such as GANs. Note that there is also the Latent Diffusion Model (LDM) where the diffusion is performed in the smaller latent space instead of the image space; this change significantly lowers the computational cost as the dimensionality of the transformed noise distribution is reduced remarkably \cite{rombach2022high}. In a similar yet different approach called Score-based Generative Models(SGM), while perturbing the image in the forward chain, we also estimate a score to keep the noise level, and then we condition the model with these scores to be decreased during the denoising process \cite{song2020score}. These techniques are the foundations for other diffusion-based generation models both for 2D novel view synthesis \cite{watson2022novel, gu2023nerfdiff, tseng2023consistent} and 3D domains \cite{Zhou_2023_CVPR, anciukevivcius2023renderdiffusion}.

% * 3D representation, NeuS, NeRF -Deniz
\subsection{Neural 3D Representations}
% voxels, meshes, point clouds
% DEEP Shape Representation: DeepSDF, DeepLS, 
% Deep Scene representation: SRN, NerF, VolSDF, NeuS, InstantNGP, MonoSDF, Gaussian Splatting
Traditionally 3D models and shapes have usually been represented using voxels, meshes, and point clouds, which can be classified as explicit representations; these representations do not require complex steps to be rendered. However, each of these representations brings their shortcomings such as memory complexity and detail capabilities. Therefore, instead of standard techniques, 3D shapes are represented implicitly using neural networks. DeepSDF and DeepLS are two of the 3D shape representation techniques, where the models are stored as learned signed distance functions \cite{chabra2020deep, park2019deepsdf}. This method was also applied in several other techniques such as VolSDF, InstantNGP, and MonoSDF \cite{yariv2021volume, mueller2022instant, yu2022monosdf}. On the other hand, neural radiance fields(NeRF) are also used to model 3D shapes, where the mapping between 5D positions and color-density pairs is learned implicitly by a multilayer perceptron \cite{mildenhall2020nerf}. Recently a new kind of 3D scene representation in the name of Gaussian splatting has been introduced where the 3D model is learned as a number of Gaussian distributions with tunable mean(position of the splat in the 3D space) and standard deviations(the size along the axes) \cite{kerbl20233d}. Note that Gaussian splats are also explicit representations, they can both consistently learn details and be rendered without any middle step.

\subsection{Single Image 3D Model Generation}
Most 3D model generation techniques require many views of the same scene, however, some works aim to obtain multiple views from a single image using diffusion techniques. However generating multiple views directly usually results in inconsistent 3D model outputs \cite{liu2023zero1to3, deng2023nerdi}. To overcome this issue there have been several works to make the generated novel views consistent with each other \cite{chan2023generative, long2023wonder3d, tang2023dreamgaussian, liu2023syncdreamer, shi2023mvdream, szymanowicz2023viewset}. These techniques differ in terms of both internal operations such as the types of contextual conditions and attention mechanisms, and 3D representation techniques such as NeRF, NeuS or Gaussian splats.

\section{Material and Method} \label{sec:material_method}

This section includes the details of Material, SyncDreamer, NeRF, and NeuS. 

\subsection{Material}

\begin{figure*}[h!]
    \centering
    \includegraphics[scale=0.7]{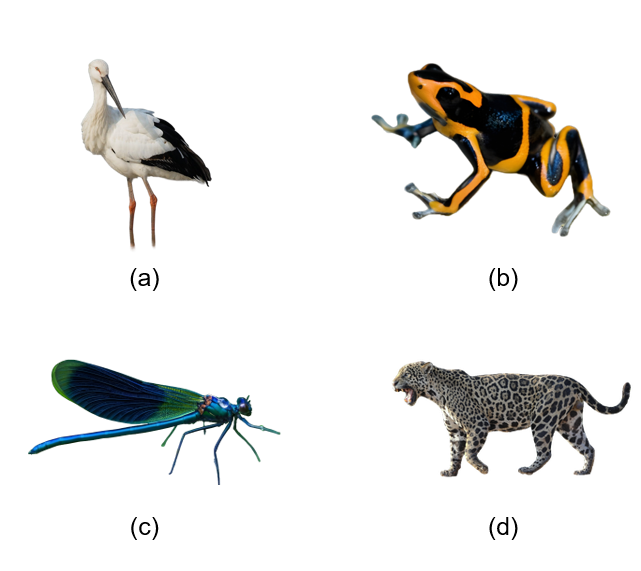}
    \caption{Sample animal images (a) oriental stork (b) frog (c) dragon fly (d) tiger.}
    \label{fig:material_images}
\end{figure*}

To monitor and conserve the health of biodiversity, we need reliable and up-to-date data. However, determining the status of all living species in the world is not an easy task. Therefore, the International Union for Conservation of Nature (IUCN) generated the Red List in 1964 \cite{iucn_redlist}. The Red List is the most comprehensive source of information that assesses and reports on the global conservation status of animal, plant, and fungal species. This information is critical to make necessary conservation decisions and trigger policy changes. The Red List is a powerful tool to protect the natural resources we need to survive. The Red List categorizes species into nine categories: Not Evaluated, Data Deficient, Least Concern, Near Threatened, Vulnerable, Endangered, Critically Endangered, Extinct in the Wild, and Extinct. These categories and criteria are an easy and widely understood system to classify species at risk of global extinction. To date, more than 150,300 species have been assessed for the Red List. The images used in this study can be seen in Fig \ref{fig:material_images}.

We are also trying to create 3D models of endangered animals to help people better understand them. By creating these models, we can study the animals in more detail and learn about their unique characteristics and behaviors. This information can then be used to develop conservation strategies that will help protect these animals and their habitats. Our goal is to raise awareness about the importance of preserving biodiversity and to inspire people to take action to protect our planet’s precious wildlife. The experiments are conducted using the SyncDreamer official implementation found in \cite{sync_official}.

\subsection{SyncDreamer}

SyncDreamer is a diffusion model that generates multiview-consistent images from a single-view image of an arbitrary object \cite{liu2023syncdreamer}. SyncDreamer models the joint probability distribution of multiview images by synchronizing the intermediate states of multiple noise predictors on every denoising step. SyncDreamer also incorporates a 3D-aware feature attention mechanism that correlates the features across different views in a 3D volume. SyncDreamer can generate images with high consistency and quality, and can be used for various 3D generation tasks such as novel-view-synthesis, text-to-3D, and image-to-3D. 

The multiview diffusion model can be formulated as follows:

\begin{equation}
p_\theta(x^{(1:N)}_0:T) = p(x^{(1:N)}_T) \prod_{t=1}^T \prod_{n=1}^N p_\theta(x^{(n)}_{t-1}|x^{(1:N)}_t)
\end{equation}

Where $x^{(n)}_t$ is the n-th target view at time step t, $p(x^{(1:N)}_T) = \mathcal{N}(x^{(1:N)}_T; 0, I)$, and $p_\theta(x^{(n)}_{t-1}|x^{(1:N)}_t) = \mathcal{N}(x^{(n)}_{t-1}; \mu^{(n)}_\theta(x^{(1:N)}_t, t), \sigma^2_t I)$. The mean $\mu^{(n)}_\theta$ is defined as:

\begin{equation}
\mu^{(n)}_\theta(x^{(1:N)}_t, t) = \frac{1}{\sqrt{\alpha_t}} x^{(n)}_t - \beta_t \sqrt{1 - \bar{\alpha}_t} \epsilon^{(n)}_\theta(x^{(1:N)}_t, t)    
\end{equation}

Where $\epsilon^{(n)}_\theta$ is the noise predictor for the n-th view, and $\alpha_t$, $\bar{\alpha}_t$, and $\beta_t$ are constants derived from the forward process. The noise predictor $\epsilon^{(n)}_\theta$ is implemented by a UNet initialized from Zero123 \cite{liu2023zero1to3} and a 3D-aware feature attention module that extracts features from a 3D volume constructed from all the target views. SyncDreamer also introduces a CLIP \cite{clip_paper} text attention layer to process the viewpoint difference and the input view as additional conditions for the noise predictor.

\subsection{NeRF}

NeRf is a technique that creates new perspectives of intricate environments through the optimization of a neural radiance field. This field is a continuous function that maps a 3D location and a 2D viewing direction to a volume density and a color that is dependent on the view. The function is approximated using a multilayer perceptron network, and its parameters are optimized by minimizing the difference between the images that are rendered and those that are observed. To achieve this, a differentiable rendering process is employed, which involves sampling 5D coordinates along camera rays and combining the corresponding colors and densities into an image, using classical volume rendering techniques.

The expected color of a camera ray $r(t) = o + td$ with near and far bounds $t_n$ and $t_f$ is:
\begin{equation}
C(r) = \int_{t_n}^{t_f} T(t)\sigma(r(t))c(r(t), d)dt 
\end{equation}

$$
\quad T(t) = \exp\left(-\int_{t_n}^t \sigma(r(s))ds\right)
$$

The color is estimated using quadrature with stratified sampling:

\begin{equation}
\hat{C}(r) = \sum_{i=1}^N T_i(1 - \exp(-\sigma_i\delta_i))c_i
\end{equation}

$$
\quad T_i = \exp\left(-\sum_{j=1}^{i-1} \sigma_j \delta_j\right)
$$

The loss function is:

\begin{equation}
L = \sum_{r\in R} \left(\hat{C}_c(r) - C(r)\right)^2 + \left(\hat{C}_f(r) - C(r)\right)^2
\end{equation}

The positional encoding function is:

\begin{equation}
\gamma(p) = \left[\sin(2^0\pi p), \cos(2^0\pi p), \sin(2^1\pi p), \cos(2^1\pi p), \dots, \sin(2^L\pi p), \cos(2^L\pi p)\right]
\end{equation}

\subsection{NeuS}

NeuS is a novel neural surface reconstruction method that learns an implicit signed distance function (SDF) from multi-view images. Unlike existing techniques that use surface rendering or volume rendering to train the SDF network, NeuS proposes a new volume rendering scheme that is unbiased and occlusion-aware in the first-order approximation of SDF. NeuS introduces a probability density function $\phi_s(f(x))$, called S-density, where $f(x)$ is the SDF and $\phi_s(x)$ is the logistic density distribution. NeuS defines a weight function $w(t)$ for each point along a camera ray $p(t)$ based on the S-density, and accumulates the colors along the ray by

\begin{equation}
C(o, v) = \int_0^\infty w(t)c(p(t), v)dt
\end{equation}

where $C(o, v)$ is the output color, $c(p(t), v)$ is the color function, and $o$ and $v$ are the camera center and direction, respectively. NeuS derives the weight function $w(t)$ by

\begin{equation}
w(t) = T(t)\rho(t)
\end{equation}

$$
\quad T(t) = \exp\left(-\int_0^t \rho(u)du\right)
$$

where $T(t)$ is the accumulated transmittance and $\rho(t)$ is the opaque density function, which is defined by

\begin{equation}
\rho(t) = \max\left(\frac{-\phi_s'(f(p(t)))}{\phi_s(f(p(t)))}, 0\right)
\end{equation}

NeuS shows that this weight function satisfies the requirements of being unbiased and occlusion-aware, and can handle complex objects with severe self-occlusions and thin structures. 

\subsection{NeRF vs NeuS}

NeRF and NeuS are two neural rendering techniques that can create novel views of complex scenes from multi-view images. NeRF uses a continuous 5D function to map a 3D location and a 2D viewing direction to a volume density and a view-dependent color. NeuS, on the other hand, learns an implicit signed distance function (SDF) from multi-view images and represents a surface as the zero-level set of the SDF. NeuS proposes a new volume rendering scheme that is unbiased and occlusion-aware in the first-order approximation of SDF.

NeuS introduces a probability density function $\phi_s(f(x))$, called S-density, where $f(x)$ is the SDF and $\phi_s(x)$ is the logistic density distribution. NeuS defines a weight function $w(t)$ for each point along a camera ray $p(t)$ based on the S-density, and accumulates the colors along the ray by

\begin{equation}
C(o, v) = \int_0^\infty w(t)c(p(t), v)dt
\end{equation}

where $C(o, v)$ is the output color, $c(p(t), v)$ is the color function, and $o$ and $v$ are the camera center and direction, respectively. NeuS derives the weight function $w(t)$ by

\begin{equation}
w(t) = T(t)\rho(t)
\end{equation}

$$
\quad T(t) = \exp\left(-\int_0^t \rho(u)du\right)
$$

Where $T(t)$ is the accumulated transmittance and $\rho(t)$ is the opaque density function.

In contrast, NeRF employs a multilayer perceptron network to estimate the 5D function and minimize the difference between the rendered and observed images by optimizing its parameters. Additionally, NeRF uses a differentiable rendering process that involves sampling 5D coordinates along camera rays and combining the corresponding colors and densities into an image, based on classical volume rendering techniques. To represent intricate, high-resolution scenes, NeRF proposes two enhancements: a positional encoding of the input coordinates to assist the multilayer perceptron in representing high-frequency functions and a hierarchical sampling approach that allocates more samples to areas with visible scene content.

Both techniques have shown promising results in synthesizing novel views of complex scenes, but they differ in their approach to neural rendering. NeuS focuses on learning an implicit SDF representation of the surface, while NeRF focuses on learning a continuous 5D function that maps a 3D location and a 2D viewing direction to a volume density and a view-dependent color. The choice of method depends on the specific application and the desired trade-off between accuracy and efficiency.

\section{Experiments and Results}

We conducted a series of experiments to evaluate the performance of SyncDreamer. In the first experiment, we generated 16 novel views of objects using SyncDreamer. The results showed that SyncDreamer was able to generate high-resolution images across different views but has several drawbacks such as maintaining consistency in geometry. In the second experiment, we trained NeRF and NeuS models using the 16 generated views. We then rendered the images using the trained model.

\begin{figure*}[h!]
    \centering
    \includegraphics[width=\linewidth]{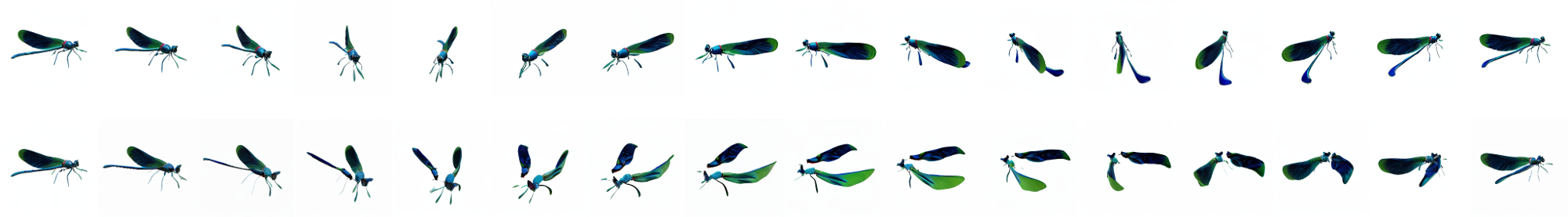}
    \caption{Novel views of Dragon Fly.}
    \label{fig:dragon_fly}
\end{figure*}

The novel views of the Red Dragon can be seen in Fig \ref{fig:dragon_fly}. Except for the last views, the novel view synthesis results of the red dragon seem reasonable. The model is able to rotate the image, allowing us to observe views that we have never seen before. In the image below, we can see that the wings gradually separate from the body and then come together again. 

\begin{figure*}[h!]
    \centering
    \includegraphics[width=\linewidth]{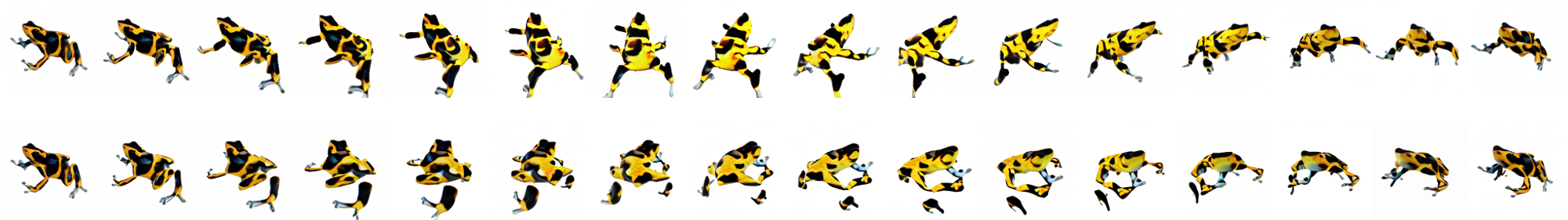}
    \caption{Novel views of Frog.}
    \label{fig:frog}
\end{figure*}

While SyncDreamer is able to generate multiview-consistent images of frogs as seen in Fig \ref{fig:frog}, it fails to maintain consistency in geometry when rotating the leg. However, the other parts of the image are generated with high quality and consistency. In the second image, a completely incorrect production has been generated.

\begin{figure*}[h!]
    \centering
    \includegraphics[width=\linewidth]{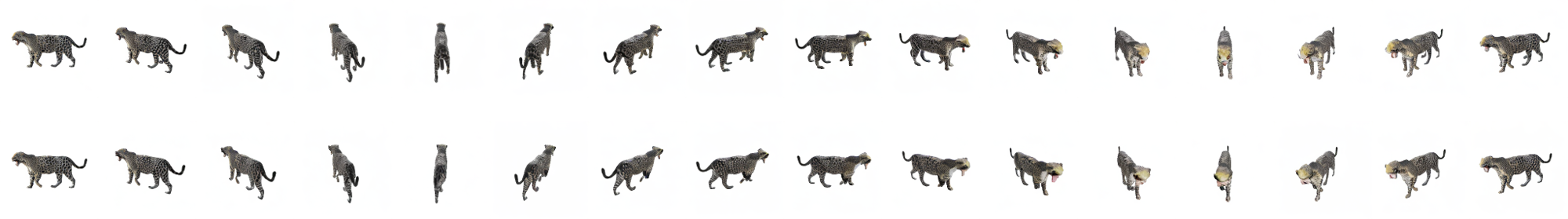}
    \caption{Novel views of Tiger.}
    \label{fig:tiger}
\end{figure*}

As shown in Fig \ref{fig:tiger}, Tigers are one of the best animals for SyncDreamer, as all of their views seem to be suitable for use. The model is able to generate images with high consistency and quality across different views, with minimal distortion. This is a remarkable achievement that shows the power of the model to generate realistic and detailed images of complex objects.

\begin{figure*}[h!]
    \centering
    \includegraphics[width=\linewidth]{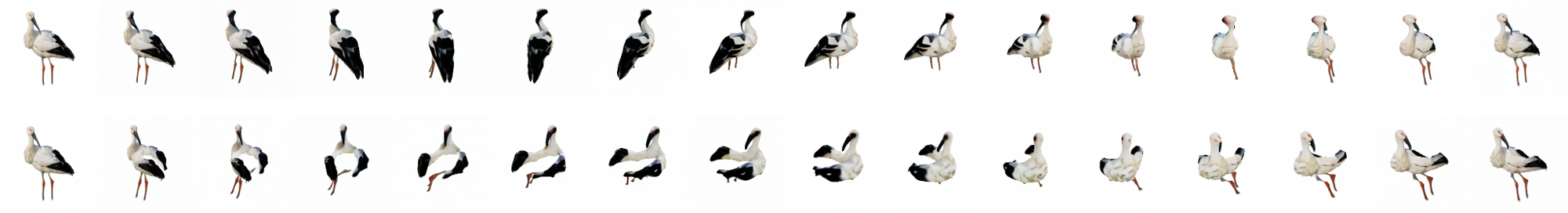}
    \caption{Novel views of Stork.}
    \label{fig:stork}
\end{figure*}

The novel views of the Oriental Stork can be seen in Fig \ref{fig:stork}. While generating images of Oriental storks, SyncDreamer fails to maintain consistency in geometry when generating the legs. Although the leg part is generated correctly in the initial positions, distortions appear in the leg and head regions towards the final positions. This is a limitation of the model that needs to be addressed in future work.

\begin{figure*}[h!]
    \centering
    \includegraphics[width=\linewidth]{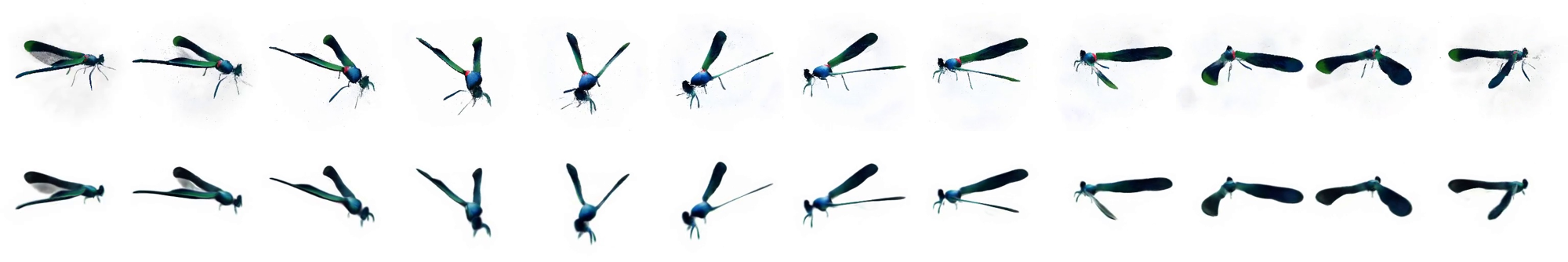}
    \caption{NeRF (first row) and NeuS (second row) outputs of Dragon Fly.}
    \label{fig:dragon3d}
\end{figure*}

\begin{figure*}[h!]
    \centering
    \includegraphics[width=\linewidth]{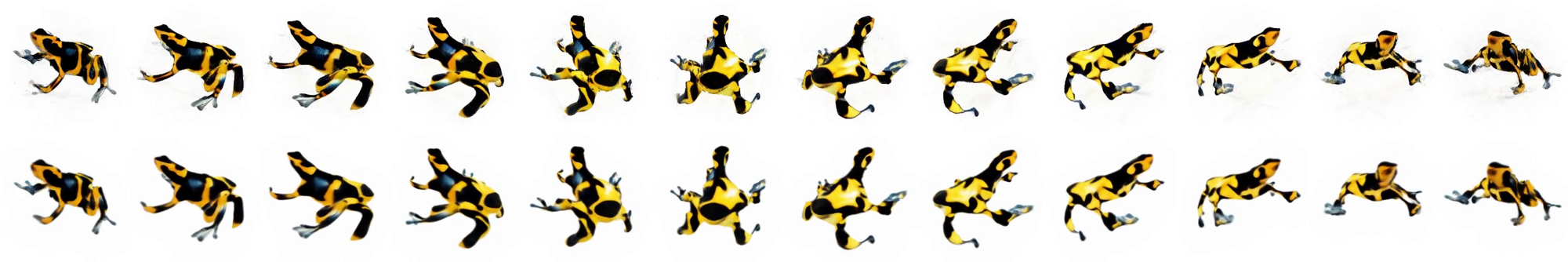}
    \caption{NeRF (first row) and NeuS (second row) outputs of Frog.}
    \label{fig:frog3d}
\end{figure*}

The NeRF and NeuS outputs are given in Fig \ref{fig:dragon3d} and Fig \ref{fig:frog3d}, respectively. When comparing Neus and Nerf outputs, it has been observed that Neus outputs are more blurry than Nerf outputs. On the other hand, Nerf outputs are noisy while Neus outputs are not. Therefore, the decision-making process can be a bit challenging. To reduce the blurriness or noise, an extra step needs to be added.

\section{Discussion}

To evaluate the performance of SyncDreamer, a series of experiments were conducted. The first experiment involved generating 16 novel views of objects using SyncDreamer. Although the results showed that SyncDreamer was able to generate high-resolution images across different views, it had some drawbacks in maintaining consistency in geometry. NeRF and a NeuS models were trained using the 16 generated views, and the images were rendered using the trained model. 

The red dragon’s novel view synthesis results are remarkable, except for the last views. The model can rotate the image, allowing us to observe views that we have never seen before. In the image below, we can see that the wings gradually separate from the body and then come together again. 

Frogs can be generated with multiview-consistent images using SyncDreamer, but the model fails to maintain consistency in geometry when rotating the leg. However, the other parts of the image are generated with high quality and consistency. In the second image, a completely incorrect production has been generated.

Tigers are one of the best animals for SyncDreamer, as all of their views seem to be suitable for use. The model can generate images with high consistency and quality across different views, with minimal distortion. This is a remarkable achievement that demonstrates the power of the model to generate realistic and detailed images of complex objects.

When comparing Neus and Nerf outputs, it has been observed that Neus outputs are more blurry than Nerf outputs. On the other hand, Nerf outputs are noisy while Neus outputs are not. Therefore, the decision-making process can be a bit challenging. To reduce the blurriness or noise, an extra step needs to be added. 

\section{Conclusion}

3D reconstruction is one of the most important applications of our time. It h
as a wide range of applications in various fields such as computer vision, robotics, entertainment, reverse engineering, augmented reality, human-computer interaction, and animation. By using a single RGB image, it is possible to create images of it from different angles and create 3D models of objects with these images. 

Studies on endangered species will help these animals to be better understood by humans. Using SyncDreamer, novel view synthesis outputs of images were obtained, and using NeuS and NerF, these output images were converted into 3D objects. SyncDreamer generates multiview-consistent images from a single-view image, while NeuS and NerF generate high-resolution 3D models from 2D images. Although some distortions occurred in the results obtained, very good results can be obtained with the corrections to be made. 
s
Overall, the experiments conducted in the paper show the effectiveness of SyncDreamer in generating multiview-consistent images and accurate 3D models of objects. The results of the experiments suggest that SyncDreamer can be used for various 3D generation tasks such as novel-view-synthesis, text-to-3D, and image-to-3D. In future studies, different techniques are planned to be used to obtain better 3D models.

\section*{Acknowledgments}

This study was supported by Huawei Türkiye R\&D Center. 

%Bibliography
\bibliographystyle{unsrt}  
\bibliography{references}

\end{document}